\documentclass[twocolumn]{webofc}
\usepackage{float}
\usepackage[varg]{txfonts}   
\usepackage{algorithm}
\usepackage{algpseudocode}
\usepackage{multicol}
\usepackage{graphicx}
\graphicspath{{graphics/}{graphics/arch/}{Graphics/}{./}}
\begin{document}
\title{Deep Q-Network for Stochastic Process Environments}

%
%

\author{\firstname{Kuangheng "Gerald"} \lastname{He} \\Project Repository: \texttt{https://github.com/skylinehk/flappy-bird-stochastic-process-stock-predict}}

\abstract{%
  Reinforcement learning is a powerful approach for training an optimal policy to solve complex problems in a given system. This project aims to demonstrate the application of reinforcement learning in stochastic process environments with missing information, using Flappy Bird and a newly developed stock trading environment as case studies. We evaluate various structures of Deep Q-learning networks and identify the most suitable variant for the stochastic process environment. Additionally, we discuss the current challenges and propose potential improvements for further work in environment-building and reinforcement learning techniques. 
}
\maketitle
\section*{Keywords:}\label{sec:keywords}
\textbf{
Reinforcement Learning, Deep Q-Learning, ~~~~Stochastic Process, Flappy Bird, \\ Portfolio Management Strategy.
}
\section{INTRODUCTION}\label{sec:INTRODUCTION}
\begin{figure}[H]
\centering
\includegraphics[width=4cm]{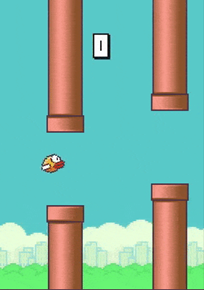}
\caption{The game image of Flappy Bird env}
\label{fig-1}
\end{figure}

The primary objective of our research is to develop optimal policies for predicting the behavior of a stochastic process environment. To achieve this, we will be using the game Flappy Bird as our starting point, as it has been previously demonstrated as a suitable environment for reinforcement learning.
In the first part of our project, our goal is to train a Deep Q-Network(DQN) agent to successfully play the game of Flappy Bird. See figure \ref{fig-1}.
The game involves navigating a bird through pipes while avoiding obstacles to earn points. The goal is to pass the bird through pipes. The observation space consists of only one variable, which is the position of the bird denoted by $s \in \mathbb{R}$. The action space includes two possible actions: moving the bird up and down denoted by $a \in {0,1}$, where 0 corresponds to moving the bird up and 1 corresponds to moving the bird down. The reward function $r(s,a)$ is defined as follows: when the bird passes through the pipe, the reward is 1, otherwise, the reward is 0. The episode terminates when the bird crashes into the pipe. The agent will be provided with position information and the current score, and it must learn to recognize the bird and pipes and locate them on its own. The game's state space is challenging, requiring the agent to generalize its learning to successfully play the game better than even human players.

 In the second part of our project, we modified the game to a stochastic process environment and developed a stock trading simulation with essential metrics. The DQN agent will be trained using a network design developed in the first part of our project and will trade using the trained policy to earn profit in a backtest class.
 
Overall, our research aims to develop a robust and generalized DQN agent that can predict and adapt to the behavior of stochastic process environments such as stock trading simulations. This will contribute to the ongoing effort to develop more efficient and effective AI algorithms that can adapt to complex and dynamic environments.
\section{RELATED WORK}\label{sec:RELATED WORK}
Previous work in this area has primarily been conducted by Google DeepMind. Mnih et al. successfully trained agents to play Atari 2600 games using deep reinforcement learning, achieving performance exceeding that of human experts on multiple games. In the domain of game environments, many successful attempts have been made to train agents using gym. For instance, Kevin Chen (2017) employed Deep Q-Networks (DQN) to train an agent to play Flappy Bird, achieving the highest score of 215.~\ref{tab-1} In the realm of stock trading, Lin Willam Cong et al. demonstrated the flexibility of using deep reinforcement learning to manage portfolios without tagging information.~
\begin{table}[H]
\centering
\caption{Highest score in flappy bird by Chen, 2017}
\label{tab-1}       
\begin{tabular}{llll}
\hline
training difficulty& flap every n& human& DQN\\\hline
easy& Inf& Inf& Inf \\
medium& 11& Inf& Inf \\
hard& 1& 65& 215 \\\hline
\end{tabular}
\end{table}

In the realm of stock trading, Lin Willam Cong et al. demonstrated the flexibility of using deep reinforcement learning to manage portfolios without tagging information.
\section{METHOD}\label{sec:METHOD}
\subsection{Deep Q-network}
\label{Deep Q-network}
The Deep Q-Network (DQN) algorithm used for the Flappy Bird environment is based on an implementation of a neural network with three linear layers and two dropout layers which takes in the number of input features, the number of hidden units, and the number of output actions as inputs. The forward function of the network applies ReLU activation to the output of each linear layer and applies dropout with a probability of 0.1 to the output of the first two linear layers. The output of the final linear layer represents the Q-values for each possible action. The DQN algorithm learns the optimal Q-values by minimizing the mean squared error loss between the predicted Q-values and the target Q-values, calculated using the Bellman equation:

\begin{equation}
Q(s, a) \leftarrow Q(s, a) + \alpha [r + \gamma \max_{a'} Q(s', a') - Q(s, a)]
\end{equation}

where $Q(s,a)$ is the Q-value for a state-action pair, $s$ is the current state, $a$ is the current action, $r$ is the reward for taking action $a$ in state $s$, $\gamma$ is the discount factor, and $\alpha$ is the learning rate. The agent selects actions according to an $\epsilon$-greedy policy, where with probability $\epsilon$ it selects a random action and with probability $1-\epsilon$ it selects the action with the highest Q-value.
\subsection{Kaiming Initialization}
\label{Kaiming initialization}
We utilized the Kaiming initialization method for the weights of our DQN network. The Kaiming initialization is a method of weight initialization to improve the convergence speed and performance of deep neural networks. This method initializes the weights using a Gaussian distribution with mean 0 and variance $\frac{2}{n}$, where $n$ is the number of input features. The Kaiming initialization is applied to the weights of the linear layers using the PyTorch with a nonlinearity of ReLU. These weight and bias initialization methods help to improve the performance of our DQN agent in learning the optimal Q-values and navigating the Flappy Bird environment.
\subsection{Replay Memory}
\label{Replay Memory}
Our implementation of the DQN algorithm in the Flappy Bird game involved the replay memory method. The ReplayMemory class stored recent transitions consisting of the current state, action taken, next state observed, and corresponding reward, in a deque data structure with a maximum capacity. The push method added a new transition to the memory buffer, and the sample method randomly selected a batch of transitions for training. This method helped decorrelate transitions, improve sample efficiency, and provide a more stable training process.
\subsection{Model Optimization}
\label{Model Optimization}
The optimize model function is realized by computing the expected state-action values using the Bellman equation defining the Huber loss function:
\begin{equation}
    \mathcal{L}(y, \hat{y}) =
    \begin{cases}
        \frac{1}{2}(y - \hat{y})^2, & \text{if } |y - \hat{y}| \leq \delta \\
        \delta|y - \hat{y}| - \frac{1}{2}\delta^2, & \text{otherwise}
    \end{cases}
\end{equation}
The function calculates the loss between expected and predicted state-action values, backpropagating the loss, clipping gradients, and performing an optimization step. It utilizes the policy and target networks to update the Q-values of the agent's policy. \\
\subsection{Flappy Bird Agent Training Pipeline}
\label{Flappy Bird Agent Training Pipeline Training Pipeline}
With the previously stated methods above, we build the full training pipeline to train the agent for Flappy bird environment in algorithm  \ref{alg:dqn}. The technique used in our approach is Q-learning with experience replay, where we store every experience at every frame in a replay memory. 
\begin{algorithm}[H]
\caption{Deep Q-learning algorithm for Flappy Bird}
\label{alg:dqn}
\begin{algorithmic}[1]
    \State Initialize DQN weights
    \State Initialize replay memory with capacity MEMORY
    \Repeat
        \State Reset the environment and observe the state
        \Repeat
            \State Compute Q-values for the current state using the policy network
            \If {random number $> \epsilon$}
                \State Take the action with the highest Q-value
            \Else
                \State Take a random action
            \EndIf
            \State Observe the transition and compute the reward
            \State Define the next state of the environment
            \State Push the transition into the replay memory
            \If {the replay memory is full}
                \State Sample a batch from the replay memory
                \State Compute the expected state-action values for the batch
                \State Compute the Huber loss between the state-action values and the expected values
                \State Zero the gradients of the optimizer
                \State Backpropagate the loss by the network
                \State Clip the gradients to prevent exploding gradients
                \State Take an optimizer step
            \EndIf
            \State Compute the soft update of the target network weights
        \Until{flappy bird crashes}
    \Until{reach max episode}
\end{algorithmic}
\end{algorithm}
\subsection{Stock Trading Environment}
\label{Stock Trading Environment}
In this part, we aim to develop a reinforcement learning algorithm to optimize stock trading decisions. To achieve this goal, we define a custom TradingEnv class to simulate stock trading in the real world. The initial state of the environment is set to a cash balance of \$ 1 million, and the commission fee is set to 0.1\%. The TradingEnv class has two methods: \texttt{reset()} and \texttt{step()}. The \texttt{reset()} method initializes the environment variables to their initial values and creates an observation consisting of four elements: the $current price$ of the stock, the $current value$ of the account, the $current position$ of the account (i.e., how many stocks the account holds), and the $current cash balance$ of the account. The \texttt{step()} method takes an action as input and updates the environment variables accordingly. The action can be to buy or sell any action value of the stock.
The current market value and balance of the account are updated after each action. The $daily profit$ and $total profit$ from the initial value of the account are calculated. A $reward$ is determined based on the total profit made by trading. The method also returns the observation, reward, and done flag. The episode is considered done when the time comes to an end, which is the last timestamp of the training data. The stock data consists of four symbolic stocks with the highest trading volume from 2018 to 2020. We will use this environment to evaluate the performance of the reinforcement learning algorithm on simulated stock trading scenarios.
\subsection{Stock Trading Agent}
\label{Stock Trading Agent}
The agent class in this project implements the deep Q-learning algorithm with experience replay to train a stock trading agent. The constructor initializes hyperparameters and memory parameters such as $\gamma$, $\epsilon$, learning rate, input dimensions, batch size, memory size, $\epsilon$ end, and $\epsilon$ decay. The action space is continuous and ranges from $-1$ to $1$. The Q-network is initialized, and the memory buffers are allocated using the input dimensions and memory size. The \texttt{store\_transition()} method stores a transition tuple of state, action, reward, next state, and done in memory. The \texttt{choose\_action()} method chooses the action with the highest value for greedy exploration or a random action for exploration based on the $\epsilon$ value. The \texttt{learn()} method computes gradients and updates the weights using a randomly generated batch index, and the Bellman equation is used to compute the Q-targets. The $\epsilon$ value is decayed over time to ensure the agent performs less exploration as it gains more experience.
\subsection{Stock Trading Agent Training Pipeline}
\label{Stock Trading Agent Training Pipeline}
With the previously stated methods above, we build the full training pipeline to train the agent for the stock trading environment in algorithm \ref{alg:training_loop_stock_trading}. 
\begin{algorithm}[H]
\caption{Training loop for stock trading environment}
\label{alg:training_loop_stock_trading}
\begin{algorithmic}[1]
\State Read training data and create a trading environment
\State Initialize the agent with hyperparameters
\State Set the number of episodes to simulate
\State Initialize empty lists for episode rewards, episode durations, and epsilon history
\For{each episode}
\State Initialize cumulative reward and done variables, and reset the environment
\While{not done}
\State Choose an action using the agent's policy
\State Take an action into the environment
\State Update the cumulative reward
\State Store the transition in the agent's memory
\State Learn from the stored transitions
\State Update the observation for the next timestep
\EndWhile
\State Append the final cumulative reward and epsilon value to the lists
\State Calculate the episode number, cumulative reward, average reward, and current epsilon
\State Append the episode duration to the list of episode durations
\State Calculate the rewards
\EndFor ~reach max episode
\end{algorithmic}
\end{algorithm}
\section{RESULTS AND DISCUSSION}
\label{RESULTS AND DISCUSSION}
\subsection{Flappy Bird Environment}
\label{Flappy Bird Environment}
\begin{figure}[H]
\centering
\includegraphics[width=5cm]{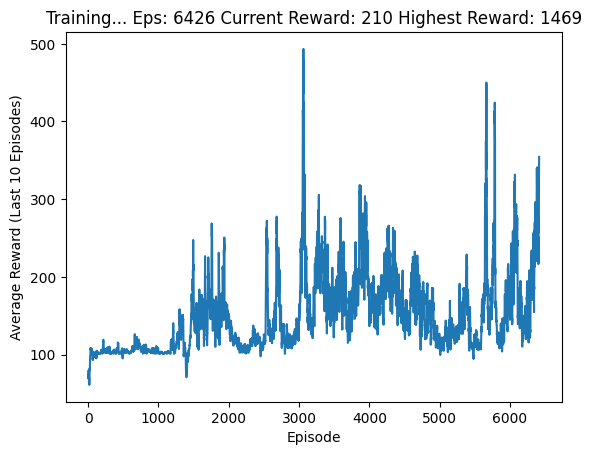}
\includegraphics[width=3cm]{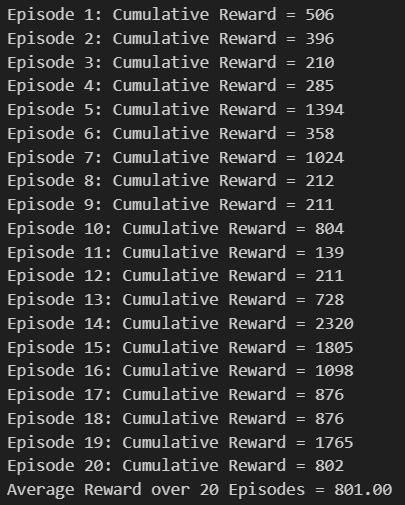}
\caption{The image of Flappy Bird training result}
\label{fig-2}
\end{figure}

We implemented an enhanced DQN for the Flappy Bird game, which incorporated dropout layers and utilized the Kaiming initialization method to improve performance. Compared to the Lunar Landing environment, where the agent receives continuous rewards, Flappy Bird only provides rewards when the agent passes through the pipes correctly. To address this, we trained a more robust network with a larger memory replay size of $MEMORY = 50000$, and a batch size of $BATCH\_SIZE = 256$, which resulted in stable activation. The enhanced DQN also employed the following hyperparameters: discount factor $\gamma = 0.99$, exploration rate starting at $\epsilon = EPS\_START = 0.99$ and decaying exponentially to $\epsilon = EPS\_END = 0.01$ over $EPS\_DECAY = 1000$ steps, target network update rate $\tau = 0.005$, learning rate $LR = 1e-4$, and hidden layer size $HIDDEN = 256$, which resulted in more stable and faster activation. The enhanced DQN triggered a faster training process and led to a significant improvement in test results. Specifically, the Average Reward over 20 Episodes increased from $158.70$ to $801.00$, while the average trigger episode decreased from $2212.73$ to $1623.58$.\ref{fig-2}.

Dropout layers randomly drop out a certain percentage of neurons during each training iteration, which prevents overfitting and encourages the network to learn more robust features. The Kaiming initialization method takes into account the non-linearity of activation functions, which helps to mitigate the vanishing/exploding gradient problem during training.\
Overall, the enhanced DQN achieved significant improvements in performance for the Flappy Bird game.
\subsection{Stock Trading Result}
\label{Stock Trading Result}
\begin{figure}[H]
\centering
\includegraphics[width=8cm,clip]{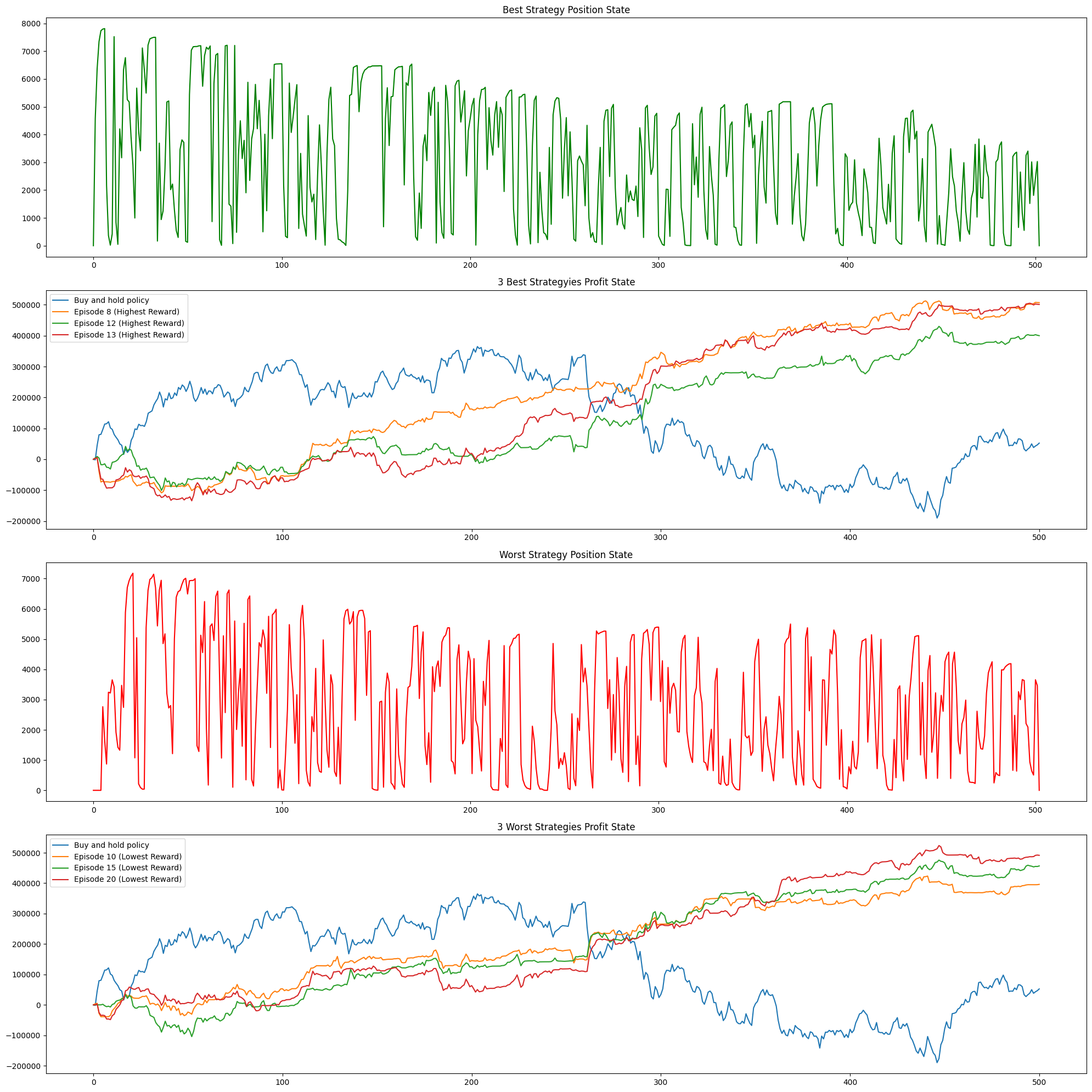}
\caption{The result plots for profit and position stats comparing with "buy and hold policy"}
\label{fig-3}       
\end{figure}

In this part, we developed a reinforcement learning algorithm to optimize stock trading decisions in both single-stock and multiple-stock environments. In the single-stock environment, the results were impressive. After training the policy for 2000 episodes, the average profit for 1000 steps was 121 \% of the starting fund. The policy tends to sell when the stock surges, mimicking the behavior of a professional investor. See fig \ref{fig-3}.
However, this is just a simulation using historical data, and the model may overfit the data. To evaluate the performance of the model, we used the backtest class to analyze the pattern of the position and profit. We found that the model's risk preference could be tuned to achieve even better results.

The backtest class allows us to analyze the performance of the policy over multiple episodes and identify patterns in the position and profit. By tuning the model's risk preference, we can further improve the performance of the policy in future simulations.
\section{CONCLUSIONS}
\label{CONCLUSIONS}
The success of our enhanced DQN in the Flappy Bird game and our previous work in stock trading demonstrate the potential of deep reinforcement learning in stochastic processes where some information is missing or incomplete. By learning directly from the pixels and score in Flappy Bird, we achieved super-human results, surpassing human-level performance. Similarly, in the stock trading environment, we were able to make successful trades despite missing or noisy data. Although training was not always consistent and overfitting or forgetting may have occurred, the potential of the DQN to successfully perform in these environments highlights the power of reinforcement learning in addressing problems where incomplete or noisy data is present. Additionally, our experience with uniform sampling from the replay memory suggests that prioritizing important experiences could lead to even better performance and more efficient training. Overall, our work underscores the promise of reinforcement learning in addressing a variety of stochastic problems and highlights several important avenues for future research in this field.

\vspace*{20cm}
%
%
%

\end{document}